\documentclass[letterpaper, 10 pt, conference]{ieeeconf}  

\IEEEoverridecommandlockouts                              

\usepackage{amsmath,amssymb,amsfonts}
\usepackage{algorithmic,algorithm}
\usepackage{xcolor}
\usepackage{url}
\usepackage{siunitx}
\usepackage{graphicx}
\usepackage{caption}
\usepackage{subcaption}
\usepackage{hyperref}
\usepackage{svg}
\usepackage{dblfloatfix}
\usepackage{multirow}
\usepackage{cite}
\usepackage{tabularray}

\usepackage{titlesec}

\usepackage{xcolor}

\titlespacing*{\section} {0pt}{5pt}{5pt}
\titlespacing*{\subsection} {0pt}{3pt}{3pt}

\graphicspath{ {./Figures/} }

\title{\LARGE \bf
Estimating Commonsense Scene Composition on Belief Scene Graphs 
}

\author{Mario A.V. Saucedo, Vignesh Kottayam Viswanathan, Christoforos Kanellakis and George Nikolakopoulos
\thanks{The authors are with Robotics \& AI Team, Department of Computer, Electrical and Space Engineering, Lule\r{a} University of Technology, Lule\r{a} SE-97187, Sweden. 
        {Corresponding author: \tt\small marval@ltu.se}}
\thanks{This work has been partially funded by the European Union's Horizon Europe Research and Innovation Programme, under the Grant Agreement No. 101119774 SPEAR.}
}%

\begin{document}

\maketitle
\thispagestyle{empty}
\pagestyle{empty}

\begin{abstract}

This work establishes the concept of commonsense scene composition, with a focus on extending Belief Scene Graphs by estimating the spatial distribution of unseen objects. Specifically, the commonsense scene composition capability refers to the understanding of the spatial relationships among related objects in the scene, which in this article is modeled as a joint probability distribution for all possible locations of the semantic object class. The proposed framework includes two variants of a Correlation Information (CECI) model for learning probability distributions: (i) a baseline approach based on a Graph Convolutional Network, and (ii) a neuro-symbolic extension that integrates a spatial ontology based on Large Language Models (LLMs). Furthermore, this article provides a detailed description of the dataset generation process for such tasks. Finally, the framework has been validated through multiple runs on simulated data, as well as in a real-world indoor environment, demonstrating its ability to spatially interpret scenes across different room types.
For a video of the article, showcasing the experimental demonstration, please refer to the following link: \url{https://youtu.be/f0tqtPVFZ2A}

\end{abstract}

\section{Introduction}

In recent years in the field of robotics, there has been a growing need to enhance robotic systems with capabilities that will enable them to understand and reason about their environments in a manner similar to humans. In this context, the capability for autonomous agents to analyze and reason over the scenes composition based on partial information of the environment is paramount when tasked with real-life applications such as exploration of subterranean environments \cite{patel2024stage}, urban inspection \cite{Viswanathan2023} or multi-modal task allocation \cite{niklas2023reactive}.

The position of objects within human-centric environments tends to follow a common distinctive trend, where the placement of most objects is strongly correlated to the functionality of the objects, and their interactions with other objects within the same space. For example, a sofa is commonly placed in front of the TV. This intuitive understanding of a scene composition allows humans to efficiently locate unseen objects based on the position of the already-seen ones. 

Currently incorporating such capabilities to robotic systems presents a series of challenges. The first one is scalability, most civic environments consist of multiple levels each one with various rooms, where the systems must use the information of the scene as a whole to match human behavior. Although predictions at a room-level are feasible~\cite{giuliari2023level}, this limits the scope of information to a substantially smaller subset that may fail to represent the overall scene composition. For example, in a school, knowing the composition of one of the classrooms highly affects our expectation for the composition of the remaining ones. 

The second challenge relates to the uncertainty of object placement in dynamic environments, where rooms structure changes according to the needs of the users. In the literature, some frameworks rely on apriori known spatial configurations for finding objects~\cite{ginting2024seek}, which sometimes in such environments can be unreliable. In human-centric environments the scene composition represents the lower unit of reasoning for further semantic classifications, in other words, is the scene composition that defines higher-level attributes like room purpose. For example, a room may be considered a bedroom or an office based on the presence of a bed.

\begin{figure*}[!htbp]
        \centering
        \includegraphics[width=\textwidth]{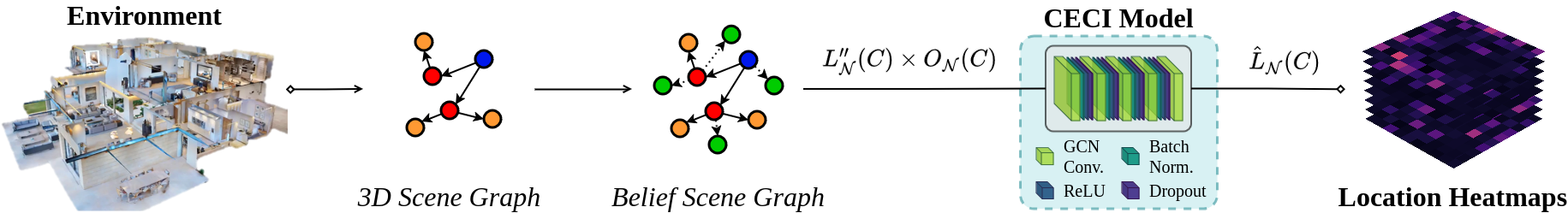}
        \caption{Depiction of the proposed method for commonsense scene composition estimation based on Belief Scene Graphs, where the BSG is an abstract representation of the scene which encompasses \textcolor[HTML]{0018EC}{building}, \textcolor[HTML]{FF0000}{rooms}, \textcolor[HTML]{FF8000}{objects} and \textcolor[HTML]{00CC00}{\textit{blind nodes}}. The graph information is then encoded as the set of location heatmaps $L^{\prime\prime}_\mathcal{N}(C) \times O_\mathcal{N}$ that are input to the proposed CECI model to output the predicted set of location heatmaps $\hat{L}_\mathcal{N}(C)$.}
        \label{fig:diag}
        \vspace{-1em}
    \end{figure*}

Recent works have tackled these challenges for the task of semantic scene completion in 3D scene graphs (3DSG)~\cite{saucedo2024belief} by levering new graph representations, namely Belief Scene Graphs (BSG), enabling task planning under uncertainty. 
Nevertheless, the estimation of node positions for members of general categories is not addressed and remains largely unexplored in the current state-of-the-art. In this work, we leverage a novel model for Computation of Expectation based on Correlation Information (CECI) for estimating commonsense scene composition on Belief Scene Graphs. An overview of our proposed approach is depicted in Fig.~\ref{fig:diag}.

\subsection*{--- Contributions}

The focus of this work is on extending a given Belief Scene Graph (BSG) with positional information about possible locations for the set of blind nodes (i.e. expected objects) in each of the rooms. Towards this end, we establish the novel concept of \textit{commonsense scene composition} as a utility-driven attribute of any given room in a BSG (Section~\ref{sec:pf}). In order to address the aforementioned challenges, we developed and implemented a Graph Convolutional Network (GCN) inspired by the Computation of Expectation based on Correlation Information (CECI) model (Section!\ref{sec:cc}). The proposed model allows to estimate the probability distribution for the position of a set of labeled semantic classes by levering the partial information of the current scene composition and the expected objects (i.e. blind nodes).

Furthermore, we provide the methodology for the generation of a dataset based on semantically annotated real-life 3D spaces, including the augmentation of the data (Section~\ref{sec:dg}) and the encoding of the location information within the nodes (Section~\ref{sec:gt}).
Additionally, we present a detailed explanation for the development of a spatial ontology using a Large Language Model (LLM) (Section~\ref{sec:so}). Using the proposed spatial ontology we develop a neuro-symbolic approach to the proposed CECI model, and present extensive evaluation of the performance of both variants (Section~\ref{sec:vm}).

Finally, we present real-world experiments on estimating commonsense scene composition by constructing a 3D scene graph using information gathered by a legged robotic platform through the exploration of an unknown indoor environment (Section~\ref{sec:ft}). The experimental results demonstrate the performance of the proposed method for estimating the commonsense scene composition of different room types.

\section{Related Work}

Currently, 3D scene graphs have emerged as a popular alternative for the encoding of scenes into intuitive and manageable abstract data representations comprising of several layers (e.g. buildings, rooms, and objects), where nodes represent concepts while edges represent relationships \cite{Armeni2019,Kim2020,Wald2020,Rosinol2021,Wu2021,Hughes2022,Wald2022,room,Feng2023,Wang2023,Wu2023}. Among the most recent works in the field we can find the introduction of belief as an enhancing capability in 3D scene graphs. 
Some of the works exploring this concept include \cite{Looper2023}, where a variable scene graph for semantic scene variability estimation is proposed. The presented framework considers three varieties of semantic scene change, such as changes in position, semantic state, or scene composition.
Similarly, the authors in \cite{ginting2023} devised a semantic belief graph aimed at navigating through environments fraught with perceptual uncertainty, offering a robust framework for semantic-based planning in extreme conditions. Furthermore, we can find \cite{saucedo2024affordance}, which proposed the concept of commonsense affordance. The focus of the work is on reasoning how to effectively identify object’s inherent utility during the task execution through the analysis of contextual relations of sparse information of 3D scene graphs.

On the other hand, in the vein of object search 
we have~\cite{giuliari2023level}, where the authors proposed an end-to-end solution to address the problem of object localization in partial scenes. 
Towards this end, a novel scene representation namely Directed Spatial Commonsense Graph (D-SCG), is described, which consists of a spatial scene graph enriched with additional concept nodes from a commonsense knowledge base. 
The proposed graph-based scene representation is then used to estimate the unknown position of the target object using a Graph Neural Network (GNN) that implements a sparse attentional message passing mechanism.

Likewise, the authors in \cite{ginting2024seek} address the problem of object-goal navigation for autonomous inspections 
by levering a Relational Semantic Network (RSN) to estimate the probability of finding the target object across spatial elements in a Dynamic Scene Graph (DSG). The introduced framework enables robots to use semantic knowledge from prior spatial configurations of the environment and semantic common sense knowledge to search for and navigate toward target objects more efficiently. Our work parallels these methods by incorporating BSGs as the scene representation, which enables the scalability of the framework and limits the effects of uncertainty on the environment.

Finally, within the contexts of semantic scene completion, we can find~\cite{saucedo2024belief}, in where a novel graph representation is presented, namely Belief Scene Graphs. In a broad sense, a BSG can be seen as an expansion of a 3D scene graph that incorporates a special kind of nodes of probabilistic nature that depict unseen objects in the environment (i.e. expected objects).
This graph representation is then used to optimize the search of objects in real-life unknown environments by narrowing the search space based on the rooms with the highest probability of containing the desired objects.

Nevertheless, in real-life environments, the presence of different perception challenges (e.g. occlusion, partial observability, etc.) hinders the efficacy that a typical autonomous robotic system would have when inspecting an area on a mission deployment, even if delimited to a single contained space (e.g. a room). 
Furthermore, in multi-modality and multi-agent scenarios, the knowledge of the room alone may lack enough insights for the assignment of specific agents to inspect or retrieve an object.

This brings forward the need for a method that levels the information about the perceived and expected objects encoded in the nodes of a BSG into an abstract representation of the scene composition. Towards this end, we proposed a GNN inspired by the CECI model architecture, which uses sparse information about the localization of the known objects along side the information of the expected objects to estimate the probability distribution for the location of the different classes present in any given BSG.

\section{Problem Formulation}
\label{sec:pf}


\textbf{Commonsense Scene Composition:} 
The position of an object is commonly correlated to the positions of the other objects that are expected to interact with it or are related to it. 
In this article, we would refer to this knowledge as commonsense scene composition $L_i^*$, which is defined as a joint probability distribution for all possible locations of the semantic class $c_i$. Let us now consider our environment to be represented by a grid \(S \times S \,|\, S \in \mathbb{Z}^+\), we have that:
\begin{equation*}
    L_i^* = \{P(c_i) \mid c_i \in C \} \quad \forall \, P : (S \times S).
\end{equation*}

\textbf{Belief Scene Graphs:} 
A Belief Scene Graph (BSG)~\cite{saucedo2024belief} is defined as a tuple of nodes and edges $\mathcal{G}^{\prime\prime} = (\{\mathcal{B}, \mathcal{V}\},\mathcal{E})$, where $\mathcal{E}$ is the set of directional edges, $\mathcal{V}$ is the set of nodes representing buildings, rooms and the observed objects, and $\mathcal{B}$ is the set of \textit{blind nodes}. Where \textit{blind nodes} are used to embody the estimated probabilities for finding objects that were not originally perceived on the respective parent node (i.e. room node). 
Let $\mathcal{G}^{\prime\prime}$ then denote a BSG with shape $(\{\mathcal{B}, \mathcal{V}\},\mathcal{E})$, and where the attributes for any given node $\nu \in \mathcal{V}^{Objects}$ are the position $p_\nu = (x, y, z) \in \mathbb{R}^3$, and the semantic class label $c_i \in C$, where $C = \{c_1, ..., c_n\}$ with $n \in \mathbb{Z^+}$ denotes the set of semantic class labels present in the BSG $\mathcal{G}^{\prime\prime}$.
In this work, we look to find a reasonable approximate $\hat{L}_i$ for the commonsense scene composition $L_i^*$ of any given semantic class $c_i$ in the BSG $\mathcal{G}^{\prime\prime}$.

\textbf{Computation of Expectation based on Correlation Information:}
Let us assume a function $\mathfrak{f}(\mathcal{G}^{\prime\prime}) = \hat{L}_C \approx L_C^*$ that approximates the commonsense scene composition of the set of semantic class labels $C$ based on the information encoded on the BSG $\mathcal{G}^{\prime\prime}$. To allow for flexibility and scalability, this function will be implemented using a GCN inspired by the Computation of Expectation based on Correlation Information (CECI) model proposed in \cite{saucedo2024belief}. 
%

\section{Commonsense Scene Composition}

This Section details the optimization and design of the proposed CECI model for commonsense scene composition estimation, as well as the methodology used to develop the dataset based on semantically annotated 3D spaces.

\subsection{CECI Model Architecture}
\label{sec:cc}

The CECI model was first introduced in \cite{saucedo2024belief}, where a novel GCN model is presented to estimate the probability distributions for finding certain objects in the different room nodes of a 3DSG. CECI models, different from traditional GCN models, are structured around the idea of estimating probability distributions rather than precise values. 
This particular behavior facilitates to estimate the commonsense scene composition as a joint probability distribution rather than precise positions, which closer resembles human knowledge and reasoning. The proposed CECI model consists of 5 GCN convolutional layers~\cite{gcn}, followed by batch normalization~\cite{batchn}, ReLU, and dropout at its base, and has been further modified to incorporate the proposed spatial ontology into a neuro-symbolic architecture (Section~\ref{sec:so}). 
The overall network architecture is visualized in Fig.~\ref{fig:diag}.

The problem is then defined in terms of the proposed CECI model architecture, where the input is any given BSG $\mathcal{G}^{\prime\prime}$ with a set of location heatmaps $L^{\prime\prime}_\mathcal{N}(C)$ and a set of object counts $O_\mathcal{N} (C)$ as node attributes over a set of labeled object classes $C$, where $L^{\prime\prime}_\mathcal{N}(C)$ is the joint probability distribution for the partial scene composition and $O_\mathcal{N} (C) \in \mathbb{Z}^+$ is the number of object nodes and \textit{blind nodes} connected to each respective node in the set of nodes of interest $\mathcal{N} = \mathcal{V}^{Room}$. The output is the set of predicted location heatmaps $\hat{L}_\mathcal{N}(C)$ which approximates $L^*_\mathcal{N}(C)$ based on the ground truth scene composition encoded in the set of location heatmaps $L_\mathcal{N}(C)$.

\subsection{Data Generation and Augmentation}
\label{sec:dg}

This article introduces, for the first time in the current state-of-the-art, the task of commonsense scene composition estimation. Hence, no datasets are currently available for the training of the proposed CECI model. Therefore, we utilized the Matterport3D Dataset \cite{Matterport3D}, currently one of the largest datasets of real-life building-scale scenes from residential, commercial, and civic spaces, as the foundation for generating the necessary training data.

Initially, we created a custom mapping by grouping and filtering the 1659 semantic categories in the dataset into a subset of 35 carefully selected semantic class labels. The selection criteria for these labels were based on three metrics: (i) their frequency in the scenes, (ii) their relevance to robotic tasks, and (iii) the relationships they hold among themselves. Additionally, we opted to consolidate the original set of room labels into a single general class label (i.e., "room") to align with the information present in a Belief Scene Graph~\cite{saucedo2024belief} and most 3DSG~\cite{Rosinol2021, Hughes2022, room}.
Next, for each of the dataset's semantically annotated 3D spaces, we constructed a ground truth 3DSG $\mathcal{G}$, 
with three node attributes: the semantic class label, the ground truth position, and the ground truth dimensions, while the edges represent descendant relations (i.e., the child node is physically contained within the parent node). 
We then applied the augmentation process described in \cite{saucedo2024belief} to each ground truth graph $\mathcal{G}$ by randomly deleting object nodes, until 25\% of the original nodes were removed, we generated a set of augmented graphs $\mathcal{G}^\prime$. This step increases dataset diversity and helps prevent overfitting.



\subsection{Ground Truth Commonsense Scene Composition}
\label{sec:gt}

Translating the intuitive concept of commonsense scene composition into a numerical format presents several challenges. As presented in Section \ref{sec:pf}, calculating these probabilities in real-world scenarios is highly difficult due to the extensive data requirements. To address this, we generate the ground truth, and any further location heatmaps, by creating an occupancy grid of size \(S \times S\) over the plane \((X, Y)\) for each class label associated with every node in the set \(\mathcal{N}\). This process results in a set of location heatmaps, denoted as \(L_\mathcal{N}(C) \in \mathbb{R}^{\mathcal{N} \times C \times S \times S}\), and a set of object counts, denoted as \(O_\mathcal{N}(C) \in \mathbb{R}^{\mathcal{N} \times C}\). The process is repeated for all augmented graphs \(\mathcal{G}^\prime\) where the resulting heatmaps represent the distribution of any given class within any room across the graphs \(\mathcal{G}^\prime\).


Using the resulting ground truth, we now seek to generate a set of Belief Scene Graphs $\mathcal{G}^{\prime\prime}$ for training. As such, we will randomly select objects encoded in each set of location heatmaps $L_\mathcal{N}(C)$ and mark them as blind nodes, subsequently, we will remove these nodes to obtain the set of input location heatmaps $L^{\prime\prime}_\mathcal{N}(C)$. Subject to the assumption that for any given BSG $O_\mathcal{N}^{\prime\prime} = O_\mathcal{N}$ we obtain $L^{\prime\prime}_\mathcal{N}(C)$ and $O_\mathcal{N}$ which will represent the attributes of any given BSG~$\mathcal{G}^{\prime\prime}$. This assumption allows us to compare the performance of the method without the influence of input noise.

\subsection{Ontology Formulation}


Unlike a 3D scene graph, which is a concrete representation of a specific environment, an ontology is independent of any particular environment observed by the robot. For instance, in an ontology, the concept of ``chair" is not tied to a specific instance of a ``chair”, whereas in a 3DSG, the concept is directly linked to a ``chair” within the scene the robot perceives. In other words, an ontology is a graph that represents all possible concepts and their relationships (e.g., that chairs are commonly found in offices), while a 3DSG models a particular environment, essentially serving as an instantiation of the spatial ontology with actual data.


Ontologies can be manually defined based on the opinions of experts or a common consensus approach, but these methods are labor-intensive and don't scale effectively for large ontologies. Contrasting with the prevalent paradigm 
where Large Language Models (LLMs) have been used to evaluate relationships between concepts, we focus on leveraging LLMs to directly create a spatial ontology representing the commonsense knowledge of the relationships among objects and their locations within a space. This ontology can then be used to add context to the learning process of the proposed model thus improving its generalization~capabilities. 

\subsection{Language-Enabled Spatial Ontology}
\label{sec:so}

A spatial ontology is a representation of spatial knowledge, which can be modeled as a graph where nodes are abstract spatial concepts and edges are spatial relations between the concepts \cite{ontology}. 
In this work, we consider a spatial ontology where each node corresponds to either low-level spatial concepts (i.e. labeled object class) $C = \{c_1, ..., c_n\}$ or high-level spatial concepts (i.e. labeled room class) $R = \{r_1, ..., r_m\}$. 
Furthermore, we only consider descendant relations, which describe where low-level concepts are expected to be located with respect to high-level concepts. For example, an edge between a low-level concept $c_i \in C$ (e.g., ``bed”) and a high-level concept $r_j \in R$ (e.g., ``bedroom”) means that $c_i$ is expected to be located within $r_j$ (e.g., a ``bed” is located in a ``bedroom”). 
This implies that the spatial ontology is a bipartite graph with $|R| \times |C|$ biadjacency matrix $\Omega$. 

%
%

We build the spatial ontology by querying the language model for each high-level concept, asking it to generate the connections that link low-level concepts to that high-level concept, \cite{ontology}. This process can be represented as:
\[
\Omega_{ij} = \text{LLM}(c_i \leftrightarrow r_j), \quad \forall \, c_i \in C, \, r_j \in R
\]
The resulting ontology includes connections that help correlate low-level concepts based on high-level concepts (e.g. a ``chair" is likely to be located in the same room as a ``table").

    \begin{figure}[!htbp]
        \centering
        \includegraphics[width=0.9\columnwidth]{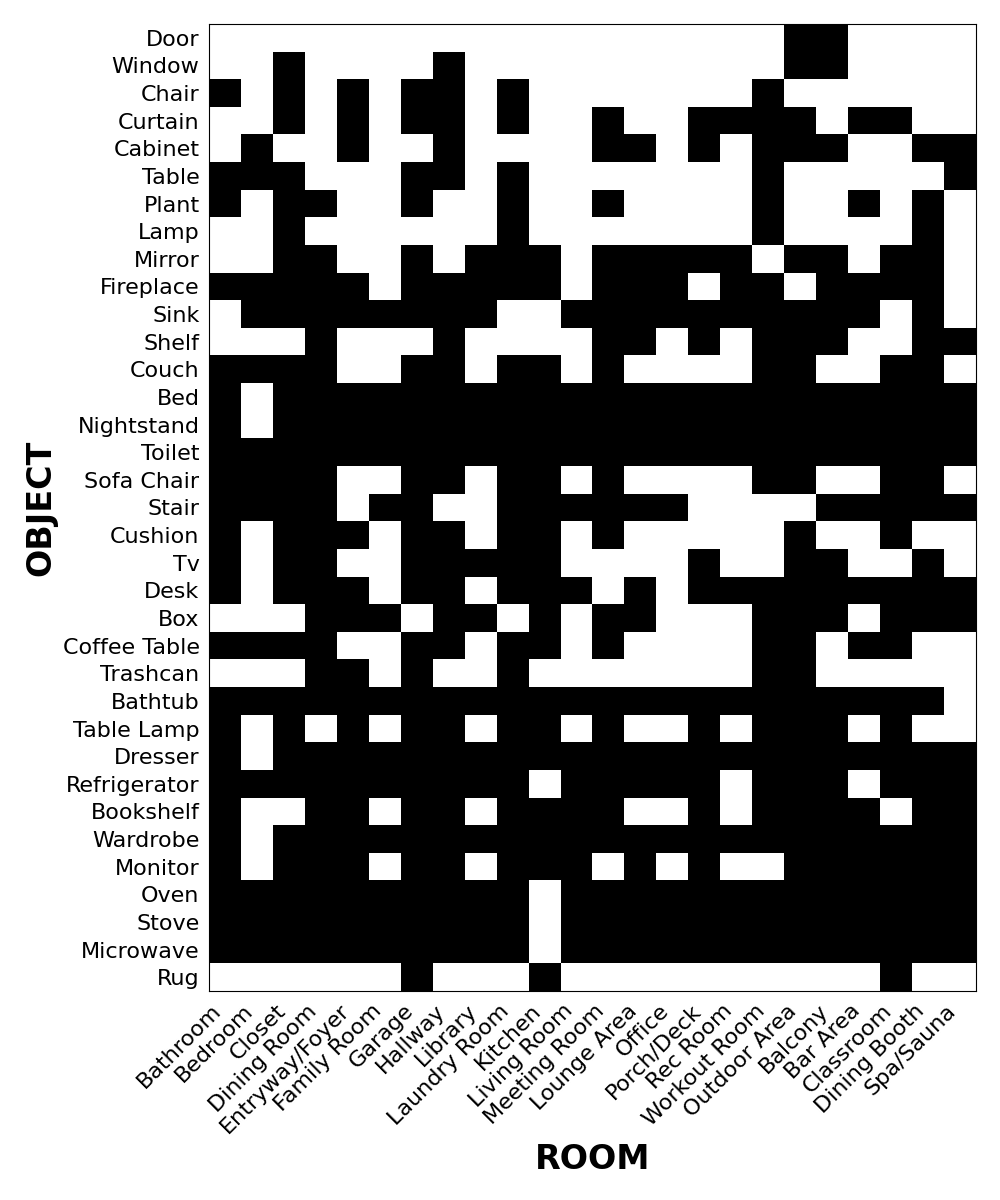}
        \caption{Language-enabled spatial ontology.}
        \label{fig:ont}
        \vspace{-1.2em}
    \end{figure}

\section{Experimental Evaluation}

In this section, we first provide the training parameters of the proposed two CECI models, i) the baseline (BASE) that uses only the information present in the input BSG and ii) the enhanced (BASE+ONT) which incorporates the proposed ontology. Secondly, we present the validation of both models using Wasserstein distance~\cite{wd}, energy distance~\cite{ed}, and Frobenius norm~\cite{fn}, followed by a qualitative analysis of the best performing model for the scene composition task. Finally, we report the results for the prediction of commonsense scene composition in a real-world indoor environment with a Boston Dynamics Spot Legged Robot.

\subsection{CECI Model Training}

In the developed implementation, we used a total of 35 labeled object classes. The generated dataset was used in an 80\% / 10\% / 10\% split for training, validation, and testing respectively.
The training consisted of 5000 epochs with a batch size of 12. We used an Adam optimizer \cite{adam} with a learning rate of 1$e^{-5}$ and a learning rate decay of 1$e^{-8}$. 
The loss function was chosen to be Mean Squared Error (MSE).

    \begin{figure*}[!h]
        \centering
        \includegraphics[width=\textwidth]{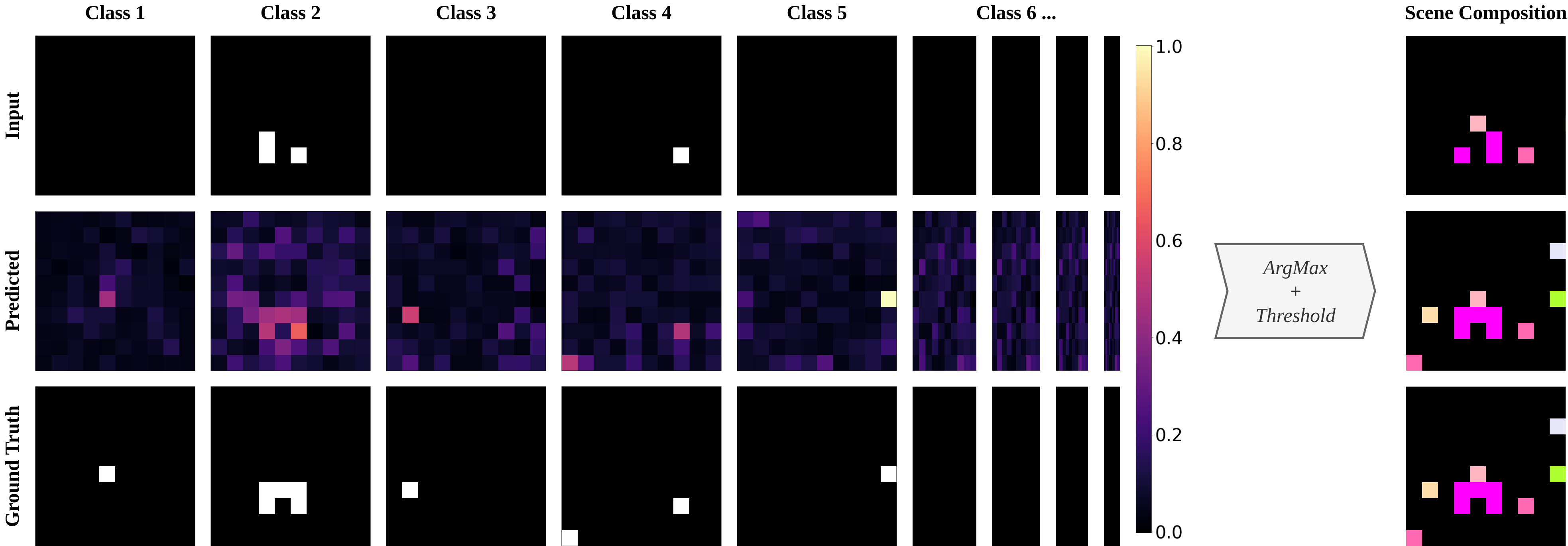}
        \caption{Depiction of the process used to construct the layout of a room based on the estimated location heatmaps for each object class. The input and ground truth of each step is also included for comparison. }
        \label{fig:res}
        \vspace{-1em}
    \end{figure*}

    \begin{figure}[!h]
        \centering
        \includegraphics[width=0.86\columnwidth]{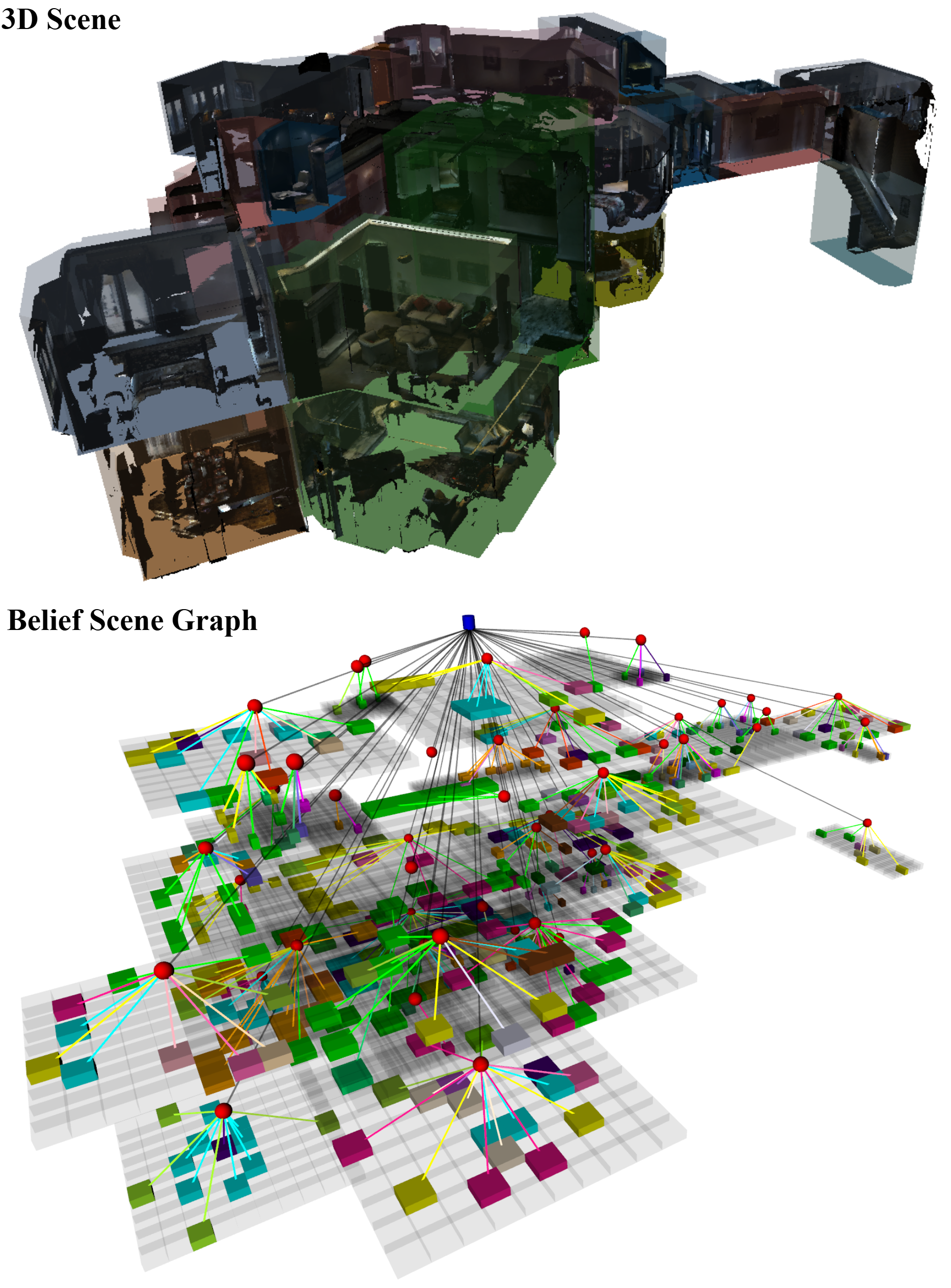}
        \caption{Depiction of the final result, where the proposed method is used to estimate the commonsense scene composition of the environment allowing to complete the Belief Scene Graph with the estimated locations. 
        }
        \label{fig:res2}
    \end{figure}

\subsection{Validation Metrics}
\label{sec:vm}

The validation process was additionally used as a ground of comparison for the performance of the proposed models in where two key metrics were evaluated.

\textbf{Statistical distance: }
The first metric was the statistical distance between the probability distributions of the predicted location heatmaps and the ground truth. This was computed using the Wasserstein distance (also known as the Earth Mover’s Distance) and the energy distance for the pair of probabilistic distributions \( \hat{L}_\mathcal{N}(C) \) and \( L_\mathcal{N}(C) \). Table~\ref{table:distances} and Table~\ref{table:distances2} show the mean, variance, skewness, and kurtosis of the Wasserstein distance and the energy distance, respectively, for each of the proposed models. Overall, the predicted distribution of both models are fairly similar to the ground truth, with the BASE+ONT model presenting slightly better performance.

\begin{table}[!htbp]
\caption{Wasserstein Distance}
\label{table:distances}
\centering
\begin{tabular}{c|c c c c}
\hline
\textbf{Metric} & \textbf{Mean} & \textbf{Variance} & \textbf{Skewness} & \textbf{Kurtosis} \\
\hline
\hline
BASE & $0.0083$ & $1.1131 \times 10^{-5}$ & $0.2297$ & $1.7656$\\
BASE+ONT & $0.0074$ & $1.3568 \times 10^{-5}$ & $0.3998$ & $-0.2638$\\
\hline
\end{tabular}
\end{table}

\begin{table}[!htbp]
\caption{Energy Distance}
\label{table:distances2}
\centering
\begin{tabular}{c|c c c c}
\hline
\textbf{Metric} & \textbf{Mean} & \textbf{Variance} & \textbf{Skewness} & \textbf{Kurtosis} \\
\hline
\hline
BASE & $0.0647$ & $9.6065 \times 10^{-5}$ & $-0.4469$ & $2.4750$\\
BASE+ONT & $0.0610$ & $1.5473 \times 10^{-4}$ & $0.1434$ & $-0.1892$\\
\hline
\end{tabular}
\vspace{-1em}
\end{table}

\textbf{Similarity: }
The second metric assessed was the similarity between the pair of probabilistic distributions \( \hat{L}_\mathcal{N}(C) \) and \( L_\mathcal{N}(C) \), measured as the Frobenius norm of the difference between the location heatmaps of the predicted and ground truth data. The calculated values in Table~\ref{table:distances3} indicate a reasonable level of similarity, reflecting the performance of the CECI model in learning the underlying correlations between different object class labels in the generated dataset. Likewise, in the previous metric, the BASE+ONT model presents slightly better performance compared to the BASE.

\begin{table}[!htbp]
\caption{Frobenius Norm}
\label{table:distances3}
\centering
\begin{tabular}{c|c c c c}
\hline
\textbf{Metric} & \textbf{Mean} & \textbf{Variance} & \textbf{Skewness} & \textbf{Kurtosis} \\
\hline
\hline
BASE & $2.0004$ & $1.5095$ & $0.8135$ & $1.1455$\\
BASE+ONT & $1.4384$ & $0.7852$ & $0.8474$ & $1.1174$\\
\hline
\end{tabular}
\vspace{-1em}
\end{table}


    
\subsection{Scene Composition}

The overall pipeline works as follows, first, we feed the proposed model with a given input BSG $G^{\prime\prime}$ containing the grid representations for each of the labeled object classes $C$, and the corresponding object counts $O(C)$, which includes the number of expected objects represented throughout blind nodes. The model will then use the learned information about the correlation of the different labeled object classes alongside the provided spatial ontology $\Omega$ to determine the commonsense scene composition of every room. This will give as a result a set of location heatmaps \( \hat{L}_\mathcal{N}(C) \) representing the probability distribution for each labeled object class within a room.
Using the resulting location heatmaps we can determine the layout of the room by computing the indices of the maximum value of all classes in each position of the grid map (i.e. ArgMax) and then setting to "empty" the location where the maximum value of all classes fails to overcome a threshold. The resulting grid is the predicted layout for the room as depicted in Fig.~\ref{fig:res}.  Finally, we can use the predicted layout to position the blind nodes within a room and such completing the scene composition of the room as depicted in Fig.~\ref{fig:res2}.

    \begin{figure*}[!ht]
        \centering
        \includegraphics[width=\textwidth]{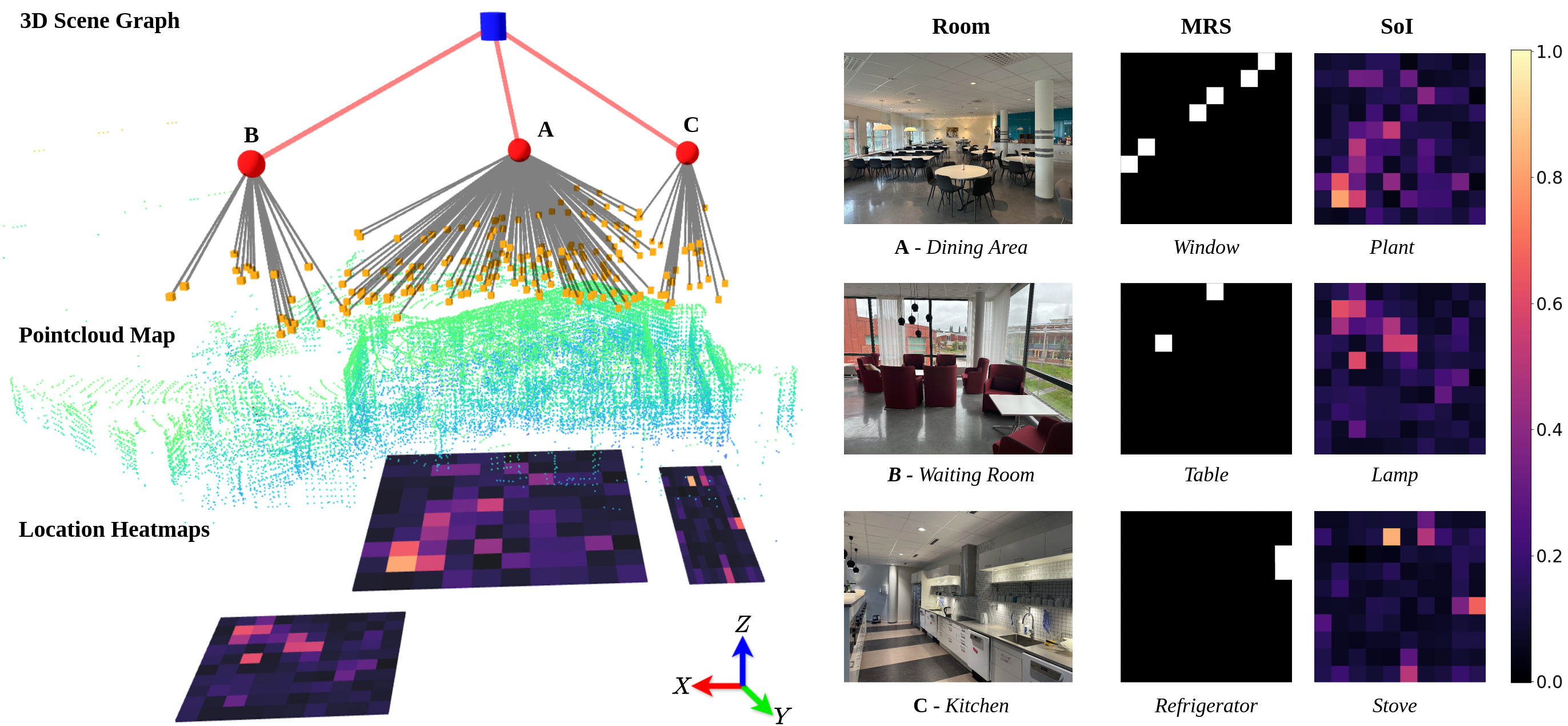}
        \caption{The generated 3D scene graph using the spot legged robot in an indoor environment, where the cylinder represents the \textcolor[HTML]{0000FF}{building}, the spheres represent \textcolor[HTML]{FF0000}{rooms}, and cubes represent \textcolor[HTML]{FF8000}{objects}. Alongside, the BSG, the scene pointcloud and the voxel representations of the estimated location heatmaps are shown. On the right, the ground truth location of the Most Relevant Semantic (MRS) for selected object classes are visualized in SoI heatmaps
        .}
        \label{fig:res3}
        \vspace{-1em}
    \end{figure*}

\subsection{Field Test Results}
\label{sec:ft}

We tested our approach on the Boston Dynamics Spot legged robot with the goal to contrast the estimated commonsense scene composition with human commonsense. 
The field experiments took place in a real civic indoor environment, which consisted of a public dining area (i.e. Room A), a small waiting room (i.e. Room B), and a shared kitchen (i.e. Room C). The overall experiment design required the robot to traverse the environment to generate a partial 3D scene graph which is later converted into a Belief Scene Graph, then enhanced using the proposed CECI model to estimate the commonsense scene composition of the environment. 


Figure \ref{fig:res3} presents the generated 3D scene graph of one of the runs, while depicting alongside the pointcloud reconstruction of the scene and the voxel representations of the estimated location heatmaps for one Semantic of Interest (SoI). The selected SoIs are among the most expected unseen-object predicted on each of the rooms, which means they were not present in the generated 3DSG and were instead estimated using a Belief Scene Graph.
Furthermore, on the right side of Fig. \ref{fig:res3}, we can find images for each of the rooms alongside the ground truth location of the Most Relevant Semantic (MRS) for the respective SoI depicted on the location heatmaps in the last column.

\textbf{Room A:} For this room the selected SoI was the object class \textit{Plant} and the MRS was the object class \textit{Window}. For this room we can observe the general behavior in where the SoI \textit{Plant} is generally placed in close proximity to the MRS \textit{Window}, being the most likely location the bottom-left corner of the room where the presence of other objects is lesser (i.e. blind nodes are generally placed in open space).

\textbf{Room B:} For this room the selected SoI was the class \textit{Lamp} and the MRS was the class \textit{Table}. Similar to the previous room, the SoI \textit{Lamp} is generally placed in close proximity to the MRS \textit{Table}, where the most likely location is at the top-left corner beside the detected MRSs.

\textbf{Room C:} For this room the selected SoI was \textit{Stove} and the MRS was \textit{Refrigerator}. Different to the previous rooms, the SoI \textit{Stove} is generally placed in mid proximity to the MRS \textit{Refrigerator}, and the most likely location is at the middle point of the walls, indicating that \textit{Stoves} are rarely found in corners or besides a \textit{Refrigerator}.

\section{Conclusions and Limitations}%

In this article, commonsense scene composition enables the extraction of information about the spatial distribution of unseen objects at the room level. This capability is critical for robotic applications, particularly in situations where systems need to predict and generate realistic scenes based on data from partially observed environments. This work proposed a CECI model to learn probability distributions throughout a GCN, while also investigated the neuro-symbolic expansion of the model with a spatial ontology based on Large Language Models (LLMs). Multiple evaluations on simulated scenes, as well as real-world field tests, showcase the applicability and performance of the framework.



One of the main limitations of the proposed method relies in the foundation of estimating scene composition based on per-room grids rather than a global one. This entails that all rooms will share an equal grid size rather than an equal voxel size as is the case with a global grid. While this design decision allows the scalability of the framework to bigger environments and the adaptability to different room sizes, it also negatively affect the precision for the localization of objects in bigger rooms. Likewise human's commonsense scene composition estimation, the predicted positions in large rooms are less accurate than those in smaller ones.

Another limitation to consider is the use of fixed size rectangle grid, which is shown in the triangle-shaped Room A. In general, the proposed method is able to learn a generalized representation of the rooms regardless of the grid shape, but it comes at the cost of computational time since the framework will still process the voxels located outside the room (e.g. top-left corner of Room A).


\addtolength{\textheight}{-6cm}   








\bibliographystyle{ieeetr}
\bibliography{References}

\end{document}